\documentclass{ar-1col}

\usepackage{amsmath, amssymb}
\usepackage{xcolor}
\usepackage{bm}

\usepackage[numbers]{natbib}
\usepackage{url}
\usepackage{soul}
\usepackage{subcaption}

\jname{Xxxx. Xxx. Xxx. Xxx.}
\jvol{AA}
\jyear{YYYY}
\doi{10.1146/(please add article doi)}

\DeclareMathOperator*{\argmin}{argmin} 

\def \StateSpace{\mathcal{X}}
\def \TotalState{X}
\def \Est{\hat{\TotalState}}
\def \MLE{\Est_{\textnormal{MLE}}}

\def \Data{\tilde{Y}}

\def \LIDS{Laboratory for Information and Decision Systems}
\def \CSAIL{Computer Science and Artificial Intelligence Laboratory}
\def \MIT{Massachusetts Institute of Technology, Cambridge, MA, USA, 02139}

\title{Advances in Inference and Representation for Simultaneous Localization and Mapping}

\author{David M. Rosen,$^1$ Kevin J. Doherty,$^2$ Antonio Ter\'an Espinoza,$^2$ and
  John J. Leonard$^2$
\affil{$^1$\LIDS, \MIT ; email: dmrosen@mit.edu}
\affil{$^2$\CSAIL, \MIT ; email: \{kdoherty, teran, jleonard\}@mit.edu}}

\usepackage[bookmarks=true, bookmarksnumbered=true, pdfpagemode=UseOutlines]{hyperref}

\begin{document}

\markboth{Rosen, Doherty, Ter\'an Espinoza, Leonard}{Advances in SLAM}

\begin{abstract}

  Simultaneous localization and mapping (SLAM) is the process of constructing a
  global model of an environment from local observations of it; this is a
  foundational capability for mobile robots, supporting such core functions as
  planning, navigation, and control. This article reviews recent progress in
  SLAM, focusing on advances in the expressive capacity of the environmental
  models used in SLAM systems (representation) and the performance of the
  algorithms used to estimate these models from data (inference). A prominent
  theme of recent SLAM research is the pursuit of environmental representations
  (including learned representations) that go beyond the classical attributes of
  geometry and appearance to model properties such as hierarchical organization,
  affordance, dynamics, and semantics; these advances equip autonomous agents
  with a more comprehensive understanding of the world, enabling more versatile
  and intelligent operation. A second major theme is a revitalized interest in
  the mathematical properties of the SLAM estimation problem itself (including
  its computational and information-theoretic performance limits); this work has
  led to the development of novel classes of certifiable and robust inference
  methods that dramatically improve the reliability of SLAM systems in
  real-world operation. We survey these advances with an emphasis on their
  ramifications for achieving robust, long-duration autonomy, and conclude with
  a discussion of open challenges and a perspective on future research
  directions.
  
\end{abstract}


\maketitle

\tableofcontents

\section{INTRODUCTION}

Simultaneous localization and mapping (SLAM) is the problem (and
procedure) of constructing a globally consistent model of an environment from
local observations of it. This is an essential capability for autonomous mobile
robots, supporting such basic functions as planning, navigation, and
control \cite{Thrun2005}. In consequence, SLAM has been the focus of an
intense and sustained research effort over the previous three decades
\cite{DurrantWhyte06ram,Bailey06ram2,Stachniss2016Simultaneous,cadena2016past}.
While this work has led to remarkable progress---including the widespread availability
of complete high-quality, open-source SLAM solutions
\cite{engel2014lsdslam,mur2017orb,whelan2015elasticfusion}---there remain
numerous fundamental challenges in the development of SLAM systems capable of
supporting truly robust, intelligent, long-duration autonomy
\cite{cadena2016past}. In this article, we survey recent progress and open
challenges in SLAM, with a particular focus on two crucial areas for achieving
this objective: inference and representation.

Inference addresses the algorithmic aspects of estimating a model of the
environment from raw sensor data. Historically, computational tractability and
empirical evaluation have been the primary drivers of algorithmic SLAM research.
This empirical orientation has enabled tremendous progress in reducing SLAM to a
practical technology, including the development of the sparse graphical
optimization-based framework that forms the basis of current state-of-the-art
techniques \cite{Stachniss2016Simultaneous,cadena2016past,Dellaert17now}. At the
same time, however, the historical emphasis on experimental evaluation (which is
restricted to measuring performance empirically, under a particular set of
operating conditions) leaves many fundamental theoretical and algorithmic
properties of the SLAM estimation problem unresolved. These include, for
example, such elementary aspects as the estimation accuracy that a SLAM system
can achieve, what specific features of a given problem determine these
performance limits, and under what conditions it is possible, even in principle,
to efficiently compute a satisfactory SLAM estimate in practice
\cite{Huang2016Critique}.\footnote{Note that SLAM estimation problems are
  typically both high-dimensional and nonconvex, which immediately raises the
  specter of computational complexity \cite{Sipser2012Introduction}.} A major
theme of recent work has been a renewed focus on these fundamental theoretical
and algorithmic challenges. In particular, we describe recent advances in
characterizing the information-theoretic limits of SLAM, the development of the
first practical inference algorithms that enjoy formal performance guarantees,
and robust extensions.

Representation concerns the environmental attributes that can be captured in a
SLAM system's model. Historically these have been grounded in simple geometric
primitives such as points, lines, and planes. A second major theme of current
research, motivated in part by recent progress in machine learning for
perception, is the development of richer representations (incorporating
properties such as temporal dynamics, objects, affordances, and semantics) to
enable higher-level reasoning and advanced autonomy, including environmental and
human-robot interaction. Until recently, such efforts were largely limited by
the necessity for a priori known object models, due to the difficulty of tasks
such as object recognition and detection. Recent years have seen researchers
revisiting seminal work such as Kuipers's spatial-semantic hierarchy
\cite{kuipers2000spatial}, now supported by decades of progress in machine
perception techniques that enable these ideas to be more fully realized. We
survey recent strides in the representational capabilities of SLAM systems,
especially the modeling and representation of nonmetric information such as
semantics and environmental topology, the ability to operate in changing
environments, the interaction of SLAM with learning, and the
increasingly task-dependent nature of representations for SLAM.

\subsection{Problem Formulation}

Before proceeding, we provide a brief review of the SLAM problem and its
mathematical formalization in order to ground our subsequent discussion. For a
more comprehensive introduction, we encourage readers to consult excellent works
by Thrun et al. \cite{Thrun2005}, Stachniss et al.
\cite{Stachniss2016Simultaneous}, and Dellaert \& Kaess \cite{Dellaert17now}.

SLAM is fundamentally the problem of constructing a consistent global model from
a collection of local observations. As real-world sensor observations are
affected by measurement noise, we formalize this problem using the language of
statistical estimation \cite{Thrun2005}. Let $ \lbrace x_i \rbrace_{i = 1}^n
\triangleq \TotalState \in \StateSpace$ denote a collection of latent states
(the model) that we would like to estimate, and  $\Data \triangleq \lbrace
\Data_k \rbrace_{k = 1}^m$ a set of sensor measurements. We assume that each
observation $\Data_k$ is sampled from a probabilistic generative model according
to:
\begin{equation}
\label{measurement_model}
\Data_k \sim p_k(\cdot | \TotalState_k ) \quad \forall k \in [m]
\end{equation}
where $\TotalState_k \subseteq \TotalState$ denotes a subset of the
states $\TotalState$ comprising the complete model.

Equation \eqref{measurement_model} formalizes our notion of locality: Each
observation $\Data_k$ depends only on a (typically very small) subset
$\TotalState_k$ of the complete model $\TotalState$. This is a ubiquitous
characteristic of SLAM problems and is a consequence of the principles of
operation of real-world sensors. For example, laser scanners and cameras provide
information only about those portions of an environment that are in a direct
line-of-sight; similarly, GPS measurements depend only on the receiver's current
position, not its previous or future locations.

The locality of the observation models in Equation \eqref{measurement_model}
enables us to decompose the joint likelihood $p(\Data | \TotalState)$ for the
model $\TotalState$ given the data $\Data$ into a product of small conditional
likelihoods:
\begin{equation}
\label{SLAM_master_eq}
 p(\Data | \TotalState) = \prod_{k = 1}^m p_k(\Data_k | \TotalState_k).
\end{equation}
This conditional factorization provides the mathematical basis for fusing the
local observations $\Data_k$ into the coherent global representation
$\TotalState$ we wish to obtain.\footnote{We remark that while Equations
  \eqref{measurement_model} and \eqref{SLAM_master_eq} as written describe the
  conditional likelihood $p(\Data | \TotalState)$ for the data $\Data$ given the
  model parameters $\TotalState$, our discussion straightforwardly extends to
  the joint distribution $p(\Data, \TotalState) = p(\Data | \TotalState)
  p(\TotalState)$ simply by appending the factor(s) describing the prior
  $p(\TotalState) = \prod_l p_l (\TotalState_l)$ to the decomposition in
  Equation \eqref{SLAM_master_eq}. Thus Equation \eqref{SLAM_master_eq} suffices
  to describe both likelihood-based and fully Bayesian formulations of the SLAM
  problem.}

\begin{figure}
\includegraphics[width=.9\columnwidth]{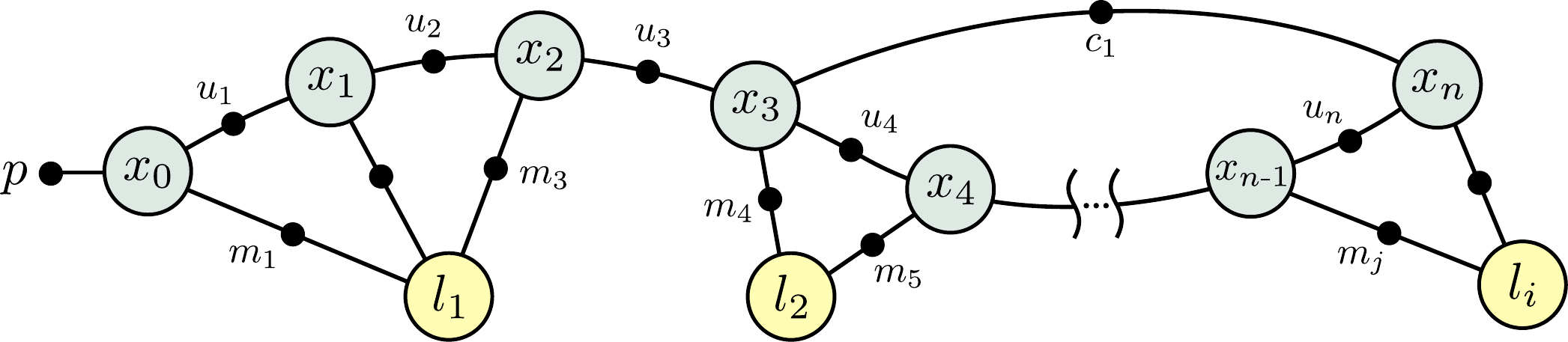}
\caption{Factor graph model of the conditional factorization shown in Equation
  \eqref{SLAM_master_eq} for a simple pose-and-landmark SLAM problem
  \cite{Rosen2014RISE}. Here variable nodes (corresponding to model parameters
  $\TotalState$) are shown as large circles, and factor nodes (corresponding to
  conditional densities $p_k$) are shown as small circles. Edges connect each
  conditional density $p_k(\Data_k | \TotalState_k)$ to the subset of variables
  $\TotalState_k$ on which it depends \cite{Dellaert17now}. In this case, the
  variables consist of robot poses $x$ and landmark positions $l$, and the
  factors are odometry measurements $u$, a prior $p$ on the initial robot pose
  $x_0$, loop closure observations $c$, and landmark measurements
  $m$. \label{fig:slam-example}}
\end{figure}

It is often convenient to model the factorization shown in Equation
\eqref{SLAM_master_eq} by means of probabilistic graphical models
\cite{Koller2009Probabilistic}. The utility of this is twofold. First, the
exploitation of the conditional independencies implied by Equation
\eqref{SLAM_master_eq} is essential for achieving fast inference, and
probabilistic graphical models make this independence structure directly
accessible via their edge sets. Second, the graphical formalism provides a
convenient modular modeling language for constructing the complex joint
distributions in Equation \eqref{SLAM_master_eq} from simple constituent parts
\cite{Dellaert17now} (Figure\ \ref{fig:slam-example}).

The probabilistic graphical model formulation described in Equations
\eqref{measurement_model} and \eqref{SLAM_master_eq} and Figure
\ref{fig:slam-example} thus provides an elegant general abstraction for the
problem of global estimation from local measurements. In particular, to
instantiate a concrete estimation problem, it suffices to specify a
representation for the model (i.e., the number and types of the states
$\TotalState \in \StateSpace$ to be estimated), and the measurement models in
Equation \eqref{measurement_model} that relate these quantities of interest to
the available data $\Data$. SLAM systems can thus be understood in terms of two
fundamental properties:

\begin{itemize}
\item \textbf{Representation:} What quantities of interest $\TotalState$ does
  the system model, and what are the generative models \eqref{measurement_model}
  that relate these parameters to the available data $\Data$?
\item \textbf{Inference:} What procedures are employed to perform inference over
  the probability distribution described by Equation \eqref{SLAM_master_eq}?
\end{itemize}

\subsection{Relation to prior surveys and scope}

The SLAM literature is vast, and a comprehensive summary of prior work is far
beyond the scope of this article. Our focus is primarily on the advances that
have occurred since 2016, when three previous surveys
\cite{cadena2016past,Huang2016Critique,lowry2016visual} were published. Readers
are encouraged to consult the standard reference by Thrun et al.~\cite{Thrun2005} for an elementary exposition of the SLAM problem, and the
tutorials by Durrant-Whyte \& Bailey~\cite{DurrantWhyte06ram,Bailey06ram2},
Stachniss et al.~\cite{Stachniss2016Simultaneous}, Dellaert \& Kaess~
\cite{Dellaert17now}, and Grisetti et al.~\cite{Grisetti2010Tutorial} for an
overview of prior algorithmic progress in SLAM.

The survey by Cadena et al.\ \cite{cadena2016past} provides an extensive
overview of the state of the art in SLAM as of 2016, including issues of
robustness, scalability, higher-level representations, and active SLAM; we
revisit many of these issues throughout this review, highlighting both recent
progress and remaining open challenges. Huang \& Dissanayake's
\cite{Huang2016Critique} critique considered algorithmic and
information-theoretic aspects of the SLAM estimation problem, including
observability, convergence, accuracy, and consistency, and is closely aligned
with our discussion of inference methods. Lowry et al.~\cite{lowry2016visual}
addressed the specific problem of place recognition, which is closely related to
issues of environmental representation and semantic mapping but aimed primarily
at the problem of identifying loop closures.

By 2016, a consensus had emerged in the research community that a certain class
of SLAM problems had become relatively well understood. For problems involving
the estimation of simple geometric primitives (such as points, lines, planes, or
camera calibrations) with well-characterized measurement models in Equation
\eqref{measurement_model} (e.g.,\ the projective mappings of vision sensors),
maximum likelihood or maximum a posteriori estimation methods built atop factor
graph representations of Equation \eqref{SLAM_master_eq} had become the method
of choice \cite{Stachniss2016Simultaneous, cadena2016past}. A number of
high-quality open-source implementations of these methods became available,
including iSAM \cite{Kaess2008iSAM}, GTSAM
\cite{Dellaert17now,Kaess2012iSAM2ijrr}, and g$^2$o \cite{Kuemmerle2011g20}, and
the availability of benchmark data sets facilitated standardized measures of
performance and steady progress. Dense and semidense visual SLAM methods,
including ORB-SLAM~\cite{mur2017orb} and ElasticFusion
\cite{whelan2015elasticfusion}, demonstrated remarkable progress in camera pose
estimation and 3D scene reconstruction for moderately sized scenes.
Visual-inertial navigation, which seeks to estimate the trajectory of a moving
sensor as accurately as possible via dead-reckoning, had also seen
substantial progress; Huang~\cite{huang2019visual} provided a recent survey of
progress in this area.

Yet despite this remarkable progress, numerous open challenges in SLAM remain,
especially regarding the representational richness and algorithmic reliability
necessary to achieve persistent, intelligent, long-duration autonomy.
Section~\ref{Geometric_estimation_section} addresses the SLAM inference problem,
including fundamental computational and information-theoretic limits,
certifiably correct estimation methods, and robust and scalable solvers.
Section~\ref{Representation_section} addresses the issue of representation for
SLAM, seeking to bring us beyond elementary geometry to consider objects and a
hierarchy of spatial relations, revisiting Kuipers's seminal work on semantic
hierarchies for spatial AI. Section~\ref{Discussion} closes the review with a
discussion of open issues and prospects for future research.

\section{ADVANCES IN ESTIMATION AND INFERENCE FOR SLAM}
\label{Geometric_estimation_section}

In this section we survey recent theoretical and algorithmic advances in
inference for SLAM. While historically computational speed and empirical
evaluation have been the primary metrics for assessing progress, a major theme
of recent work has been a renewed focus on more deeply understanding the
theoretical properties of the SLAM estimation problem in Equation
\eqref{SLAM_master_eq}, especially its geometric, algebraic, and graph-theoretic
structure. These insights have illuminated many fundamental but previously only
poorly understood aspects of the problem (including limits on achievable
accuracy and computational cost), as well as enabled the development of novel
classes of inference algorithms, including the first practical algorithms with
formal performance guarantees for nonconvex SLAM estimation problems.

\subsection{Computational hardness and the problem of nonconvexity}

The fundamental algorithmic challenge of SLAM is that the model in Equation
\eqref{SLAM_master_eq} is a high-dimensional distribution over a nonconvex state
space $\StateSpace$, and therefore performing inference within this model is
computationally hard in general \cite{Koller2009Probabilistic}. Early research
in SLAM explored a variety of approaches for performing tractable approximate
inference (using, for example, extended Kalman filters or Monte Carlo sampling)
\cite{Thrun2005}; however, by 2016 the community had settled on maximum
likelihood estimation [or more generally M-estimation \cite{Huber2004Robust}] as
the de facto method of choice \cite{cadena2016past,Dellaert17now}. In brief,
this approach recovers a point estimate $\MLE \in \StateSpace$ of the latent
state $\TotalState$ as the minimizer of an optimization problem of the form:
\begin{equation}
\label{MLE_formulation_of_SLAM}
\MLE(\Data) \triangleq  \argmin_{\TotalState \in \StateSpace} \sum_{k = 1}^m l_k(\TotalState ; \Data_k),
\end{equation}
where each summand $l_k(\TotalState_k; \Data_k)$ is the negative log-likelihood
of the corresponding factor $p_k(\Data_k | \TotalState_k)$ in the model in
Equation \eqref{SLAM_master_eq}, or a robust generalization thereof
\cite{Huber2004Robust}.

The maximum likelihood formulation in Equation \eqref{MLE_formulation_of_SLAM}
enjoys several attractive properties. From a theoretical standpoint, maximum
likelihood estimation provides strong performance guarantees on the statistical
properties of the resulting estimator (including asymptotic consistency and
normality under relatively mild conditions \cite{Ferguson1996Course}).
Computationally, the formulation of the estimation in Equation
\eqref{MLE_formulation_of_SLAM} as a sparse optimization problem admits the
application of sparsity-exploiting first- or second-order smooth optimization
methods \cite{Nocedal2006Numerical} to efficiently recover critical points of
the loss function. This computational efficiency is essential in enabling
real-time robotics applications, where both computational and temporal resources
on mobile platforms may be very limited. And indeed, current state-of-the-art
algorithms and software libraries based on the formulation in Equation
\eqref{MLE_formulation_of_SLAM} are now capable of processing SLAM problems
involving tens to hundreds of thousands of states on a single processor in real
time
\cite{Grisetti2009Nonlinear,Kuemmerle2011g20,Kaess2012iSAM2ijrr,Rosen2014RISE,williams2014concurrent}.

\begin{figure*}
\centering
\begin{minipage}[b]{.22\textwidth}
\subcaptionbox{Parking garage\label{parking_garage}}
{\includegraphics[width=\textwidth]{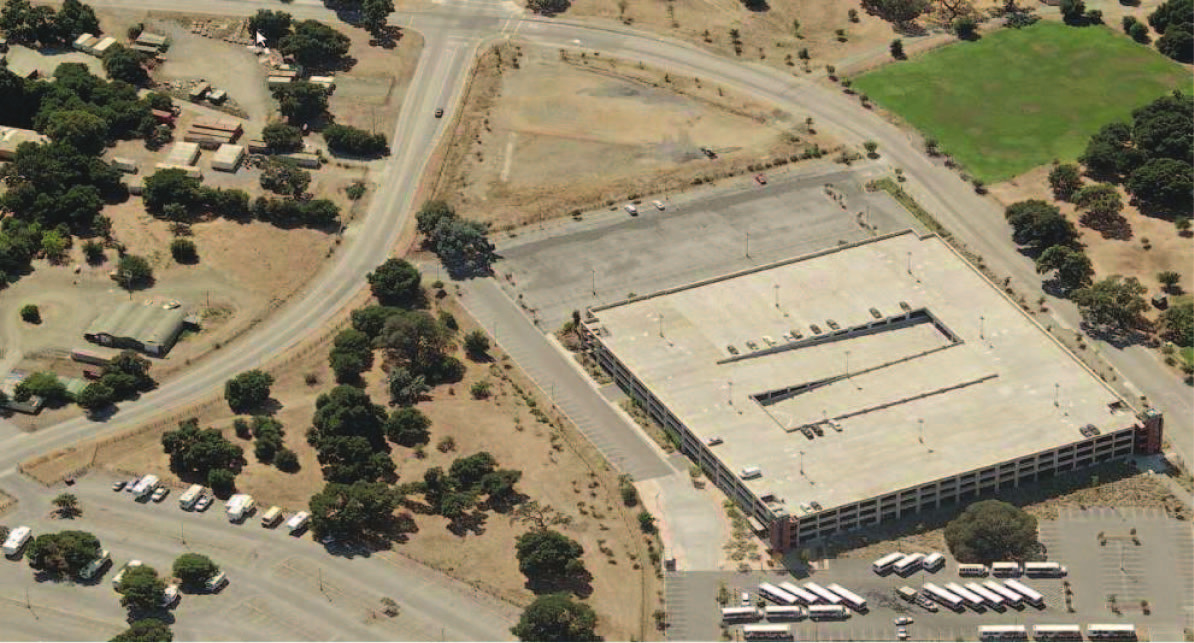}}
\end{minipage} \quad 
\begin{minipage}[b]{.22\textwidth}
\subcaptionbox{Global minimum\label{parking_garage_global_optimum}}
{\includegraphics[width=\textwidth,angle=180]{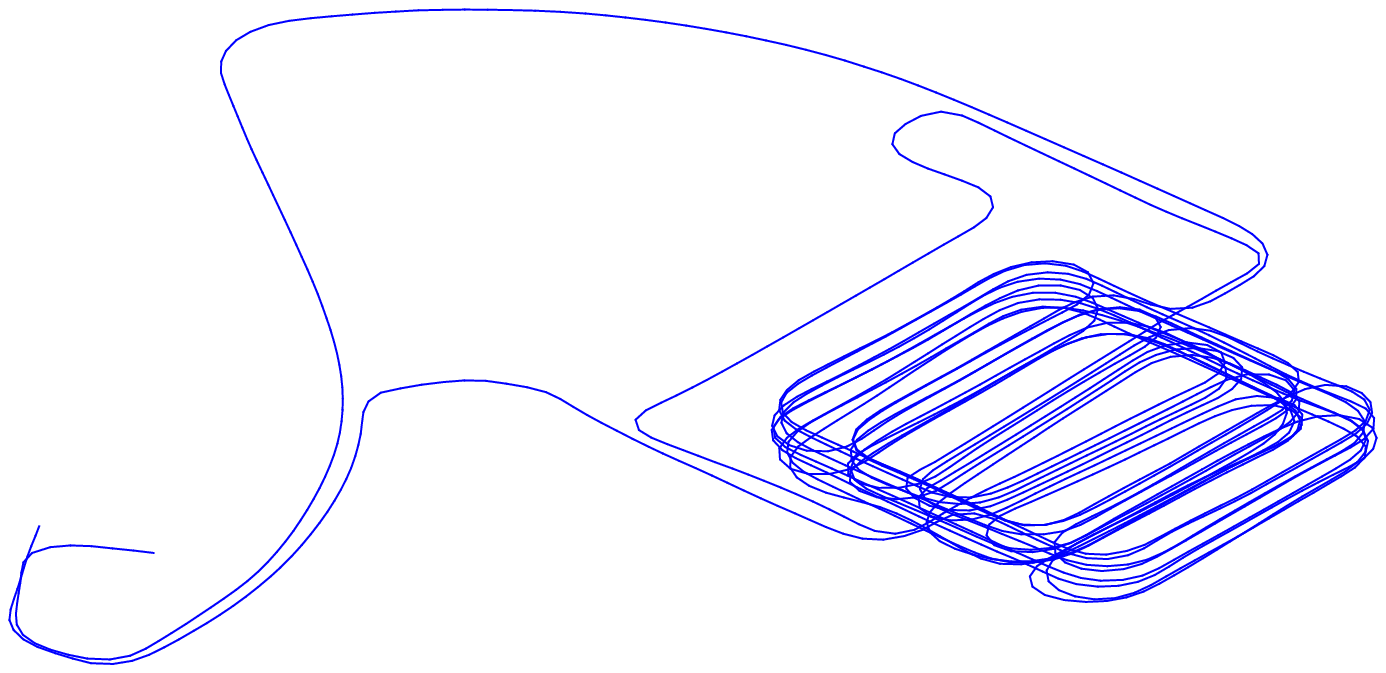}}
\end{minipage} \quad 
\begin{minipage}[b]{.22\textwidth}
\subcaptionbox{Critical point \label{parking_garage_local_minimum1}}
{\includegraphics[width=\textwidth,angle=180]{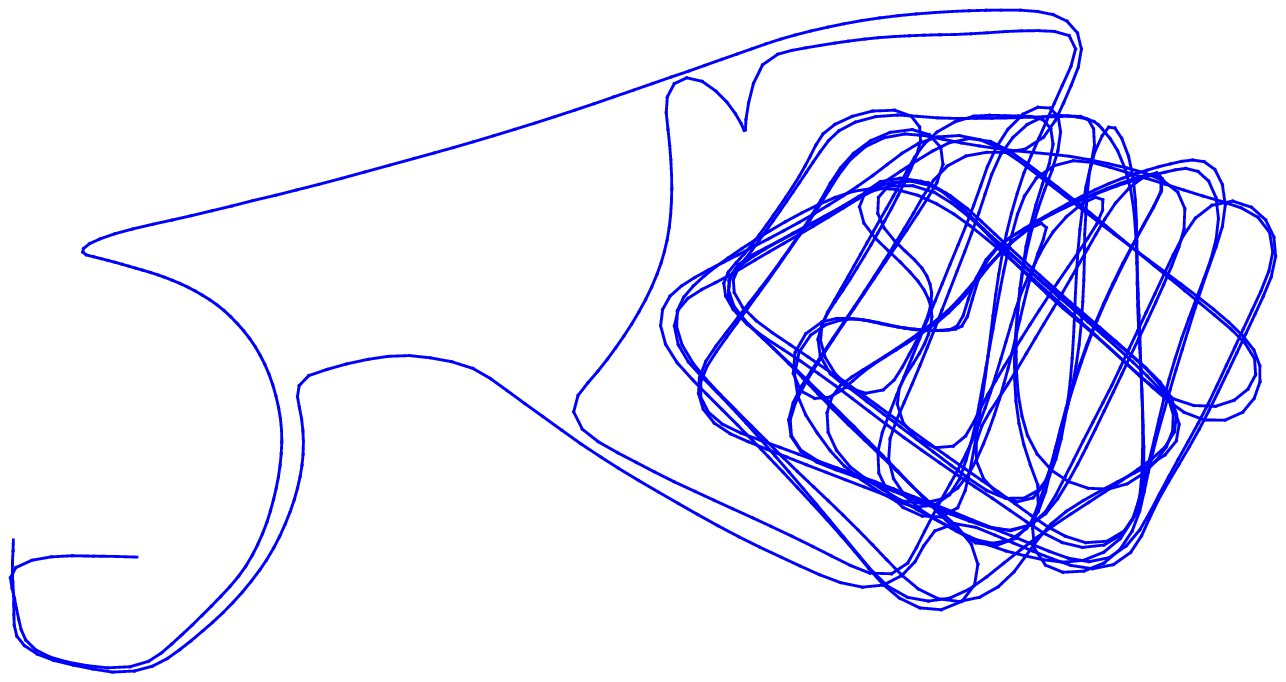}}
\end{minipage} \quad
\begin{minipage}[b]{.22\textwidth}
\subcaptionbox{Critical point \label{parking_garage_local_minimum2}}
{\includegraphics[width=\textwidth,angle=180]{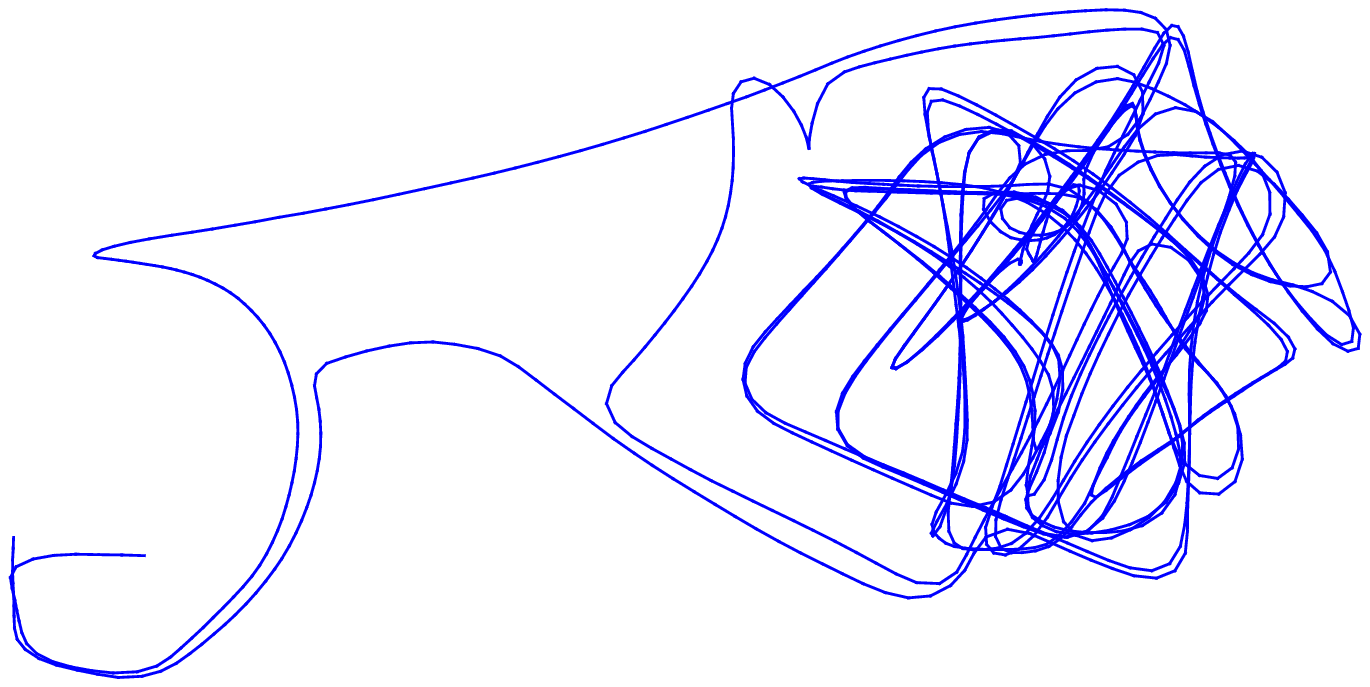}}
\end{minipage}
\vspace{\baselineskip}

\caption{Examples of suboptimal estimates in pose-graph SLAM. Several estimates
  are shown for the trajectory of a robot as it enters and explores a multilevel
  parking garage, obtained as critical points of the maximum likelihood
  estimation in Equation \eqref{MLE_formulation_of_SLAM}.
  (\subref{parking_garage}) The parking garage.
  (\subref{parking_garage_global_optimum}) The true (globally optimal) maximum
  likelihood estimate $\MLE$ computed using SE-Sync
  \cite{Rosen2016Certifiably,Rosen2019SESync}.
  (\subref{parking_garage_local_minimum1},
  \subref{parking_garage_local_minimum2}) Suboptimal critical points $\Est$
  obtained using local search. (\subref{parking_garage}) adapted from Reference
  \cite{Grisetti2010Tutorial} with permission from IEEE,
  (\subref{parking_garage_global_optimum})-(\subref{parking_garage_local_minimum2})
  adapted from Reference \cite{Rosen2019SESync}.}
\label{parking_garage_local_minima_example_fig}
\end{figure*}

However, the use of fast local optimization comes at the expense of reliability:
Local search techniques can only guarantee convergence to a critical point
$\Est$ of the loss function, rather than the global minimizer $\MLE$ required in
Equation \eqref{MLE_formulation_of_SLAM}. Moreover, it is not difficult to find
even fairly simple examples where suboptimal critical points are such poor
solutions as to be effectively unusable as SLAM estimates (Figure
\ref{parking_garage_local_minima_example_fig}). To address this potential
pitfall, several strategies for initializing local search have been proposed in
the literature, with the aim of favoring convergence to the true (global)
minimizer
\cite{Martinec2007Robust,Liu2012Convex,Carlone2015Initialization,Rosen2015Approximate,Arrigoni2016Spectral}.
While these heuristics are often effective in practice, they do not provide any
guarantees on the quality of the estimates $\Est$ that are ultimately recovered.

These algorithmic difficulties can actually be understood as particular
consequences of a fundamental computational stumbling block
\cite{Sipser2012Introduction}: As a high-dimensional nonconvex optimization, the
maximum likelihood formulation in Equation \eqref{MLE_formulation_of_SLAM} is
general enough to encompass many problems that are known to be NP-hard,
including, in particular, the fundamental problem of rotation averaging
\cite{Bandeira2016Tightness,Hartley2013Rotation}.\footnote{This also entails
  that any estimation problem in the form of Equation
  \eqref{MLE_formulation_of_SLAM} that subsumes rotation averaging---including,
  for example, the fundamental problem of pose-graph SLAM
  \cite{Rosen2019SESync}---is also NP-hard \cite{Sipser2012Introduction}.} This
implies that in fact there cannot exist an algorithm that is capable of
efficiently computing the maximum likelihood estimator $\MLE$ required in
Equation \eqref{MLE_formulation_of_SLAM} in general [unless $P = NP$
\cite{Sipser2012Introduction}].

In light of these considerations, as of 2016 several fundamental aspects of the
reliability of state-of-the-art SLAM inference methods remained poorly
understood \cite{cadena2016past,Huang2016Critique}:

\begin{itemize}
\item \textbf{Algorithmic:} Under what conditions do SLAM inference methods
  successfully recover the correct estimate $\MLE$ in Equation
  \eqref{MLE_formulation_of_SLAM}? Given the fundamental computational hardness
  of Equation \eqref{MLE_formulation_of_SLAM}, is it even possible to design
  SLAM estimation algorithms that are both practical and reliable? And if so,
  under what circumstances is this achievable?
 
\item \textbf{Statistical:} Assuming that it is possible to compute $\MLE$ in
  Equation \eqref{MLE_formulation_of_SLAM}, what are its statistical properties
  (e.g., achievable accuracy)? And what features of a given instance of Equation
  \eqref{MLE_formulation_of_SLAM} determine these properties?
\end{itemize}

\subsection{Certifiably Correct SLAM}
\label{certifiably_correct_SLAM_subsection}

One of the most exciting advances of the last few years has been the development
of the first class of SLAM estimation methods that are provably capable of
efficiently recovering optimal solutions of Equation
\eqref{MLE_formulation_of_SLAM} for nonconvex problems, at least in certain
practically important cases. These novel approaches, referred to as certifiably
correct methods, are based on employing convex relaxation, rather than smooth
local optimization, to search for high-quality state estimates. While the idea
of applying convex relaxation in SLAM is not new [indeed, many well-known
initialization techniques are based upon this strategy
\cite{Martinec2007Robust,Liu2012Convex,Carlone2015Initialization,Rosen2015Approximate,Arrigoni2016Spectral}],
what distinguishes certifiably correct methods from prior work is that the
relaxations they employ are exact\footnote{A convex relaxation is called exact
  if its minimizer provides an exact solution to the original problem from which
  it was derived. Note that this condition can be checked post hoc, simply by
  verifying that the minimizer of the relaxation satisfies the constraints of
  the original problem.} provided that the noise on the data $\Data$ in Equation
\eqref{MLE_formulation_of_SLAM} is not too large. Certifiably correct methods
thus directly tackle the fundamental problem of nonconvexity in Equation
\eqref{MLE_formulation_of_SLAM}: They enable the efficient computation of
globally optimal solutions via convex programming within a restricted (but still
practically relevant) operational regime \cite{Bandeira2016Probably}.

Certifiably correct SLAM algorithms originated in the study of pose-graph SLAM
specifically (Figure\ \ref{fig:slam-example}). Carlone and colleagues
\cite{Carlone2015Duality,Carlone2015Lagrangian, Carlone2016Planar} proposed to
address the problem of nonconvexity by leveraging Lagrangian duality. To that
end, they developed a quadratically-constrained quadratic program formulation of
the SLAM problem, and observed that the corresponding Lagrangian dual, a
semidefinite program \cite{Boyd2004Convex}, was frequently tight in practice.
This observation provides a simple means of certifying the optimality of a
(correct) candidate solution $\Est$ of Equation \eqref{MLE_formulation_of_SLAM}
using Lagrangian duality. However, due to the high computational cost of
standard (interior-point) semidefinite programming methods, this certification
approach \cite{Carlone2015Duality,Carlone2015Lagrangian} still depends on
local search in Equation \eqref{MLE_formulation_of_SLAM} to compute the candidate solution
$\Est$ itself. Rosen and colleagues \cite{Rosen2019SESync,Rosen2016Certifiably}
subsequently studied the dual of Carlone and colleagues' semidefinite relaxation
(Shor's relaxation of the original pose-graph SLAM problem), and proved that for
sufficiently small noise it admits a unique low-rank solution that provides the
exact MLE $\MLE$; they also developed a specialized low-rank Riemannian
semidefinite optimization method to efficiently solve large-scale instances
of this problem. The resulting algorithm, SE-Sync, was the first practical
certifiably correct method to appear in the SLAM literature. Rosen and
colleagues \cite{Rosen2019SESync,Rosen2016Certifiably,Rosen2019Towards} further
observed that the semidefinite relaxation and low-rank optimization methods
employed in SE-Sync could be directly generalized to a broad class of estimation
problems (specifically, those formulated as rational polynomial optimization
problems) via moment relaxation \cite{Lasserre2010Moments}, thereby providing a
general approach for synthesizing certifiably correct estimators.\footnote{Kahl
  \& Henrion \cite{Kahl2007Globally} also proposed the use of moment relaxation
  for globally optimal geometric reconstruction in computer vision, although
  they considered only low-dimensional ($\leq 11$D) estimation problems due to
  the high computational cost of standard interior-point semidefinite
  optimization methods.}

A nascent but rapidly growing body of work has subsequently adapted this
approach to produce certifiably correct estimation methods for a variety of
machine perception tasks, including rotation averaging
\cite{Eriksson2018Rotation, Dellaert2020Shonan}, calibrated two-view
registration \cite{Briales2018Certifiably}, extrinsic sensor calibration
\cite{Giamou2019Certifiably}, 3D registration \cite{Briales2017Convex,
  Yang2020Teaser}, image segmentation \cite{Hu2019Accelerated}, shape
reconstruction \cite{Yang2020Perfect}, and alternative formulations of
pose-graph SLAM \cite{Briales2017Cartan}, including sharper specialized
relaxations for the 2D case \cite{Fan2019Efficient,Mangelson2019SOS}. In recent
works, Fan \& Murphy \cite{Fan2019Proximal} and Tian et al.~\cite{Tian2019Block}
have also shown how to adapt the low-rank Riemannian optimization used by Rosen
et al. \cite{Rosen2019SESync} and Briales \& Gonzalez-Jimenez
\cite{Briales2017Cartan} to run in a distributed setting, enabling the first
distributed certifiably correct methods for pose-graph SLAM and rotation
averaging \cite{Tian2020distributed}. The further development of this class of
estimation methods remains a very active area of research.

\subsection{Robust Estimation}
\label{Certifiably_robust_subsection}

It is well known that maximum likelihood estimators are also frequently
nonrobust, meaning that corrupting an arbitrarily small fraction of the data
$\Data$ can cause the estimator $\MLE$ in Equation
\eqref{MLE_formulation_of_SLAM} to diverge from the true latent value of the
parameter $\TotalState$ \cite{Huber2004Robust}. This is true in particular for
MLEs formulated as nonlinear least squares problems, as is frequently the case
in robotics and computer vision applications. In practical SLAM applications,
corrupted (outlier) measurements of this sort frequently arise from erroneous
data associations (often due to visual aliasing) and are a primary source of the
brittleness in current state-of-the-art systems. The development of SLAM
estimation methods that are robust to outlier contamination is thus crucial for
achieving reliable, long-duration autonomy.

Several approaches for addressing the problem of outlier contamination have
previously appeared in the SLAM and computer vision literature
\cite{cadena2016past}. One line of work attempts to directly identify the set of
inlier measurements by searching for the largest subset satisfying a notion of
mutual consistency. A classical example of this class is random sample consensus
(RANSAC) \cite{Fischler1981RANSAC}, which uses random sampling to search for
sets of inlier measurements; while this approach works well for low-dimensional
problems with a moderate proportion of outliers, the curse of dimensionality
precludes its scaling to the high-dimensional problems typical in SLAM
applications. In consequence, several more scalable implementations of consensus
set search have been proposed specifically for use in SLAM. A prominent example
is realizing, reversing, recovering (RRR) \cite{Latif2013Robust}, which exploits
the sequential structure of robotics applications to iteratively construct a
consensus set: It integrates the measurements sequentially in small batches,
checking the internal consistency of the resulting model each time; any batch
that results in inconsistency is assumed to contain outliers, and is discarded.
While this approach can be very effective, it requires solving an (expensive)
high-dimensional nonlinear estimation problem each time new data are added or
removed. A more recent and computationally lightweight alternative is pairwise
consistency maximization (PCM) \cite{Mangelson2018Pairwise}; this method
constructs a graph whose vertices correspond to the measurements $\Data$ and
whose edges connect mutually consistent observations, and then extracts the
maximal clique as an estimate of the inlier set. Both RRR and PCM provide
computationally tractable consensus set estimation methods for high-dimensional
SLAM problems, and are often very effective in practice; however, they provide
no formal guarantees on the quality of the solutions they return.

An alternative class of approaches, originating in the work of Huber
\cite{Huber1964Robust}, is based on replacing the negative log-likelihood
functions $- \log p_k(\Data_k | \TotalState_k)$ that would appear in a standard
maximum likelihood formulation of Equation \eqref{MLE_formulation_of_SLAM} with robust
losses $l_k(\TotalState ; \Data_k)$ that are less sensitive to the deleterious
effects of outlier contamination; the resulting class of estimators thus
achieves improved reliability at the cost of a (typically relatively minor) loss
in statistical efficiency. Moreover, one can show that for suitable choices of
the loss function, the resulting M-estimator is provably insensitive to
contamination by a bounded fraction of outlier observations
\cite{Huber2004Robust}. Several robust M-estimation schemes have been proposed
for use in SLAM, including the well-known switchable constraints
\cite{Sunderhauf2012Switchable}, dynamic covariance scaling
\cite{Agarwal2014Dynamic}, and max-mixtures \cite{Olson2013Inference}, all of
which are straightforwardly implementable using the standard high-dimensional
local optimization machinery already prevalent in these applications
\cite{Kuemmerle2011g20,Kaess2012iSAM2ijrr,Rosen2014RISE}. Unfortunately, the
shape required of a loss function in order to attain robustness against
contamination tends to exaggerate the nonconvexity of the M-estimation in
Equation \eqref{MLE_formulation_of_SLAM}, thus rendering these techniques more
vulnerable to convergence to poor quality critical points; this pitfall is
further exacerbated by the fact that the methods typically employed to
initialize the local search
\cite{Martinec2007Robust,Carlone2015Initialization,Rosen2015Approximate,Arrigoni2016Spectral}
are themselves no longer trustworthy when faced with potentially contaminated
data.

In light of these considerations, a natural pathway to achieving practical
robust estimation is attempting to combine the robust M-estimation in Equation
\eqref{MLE_formulation_of_SLAM} with the certifiably correct global optimization
strategy outlined in Section \ref{certifiably_correct_SLAM_subsection}. To that
end, Yang \& Carlone \cite{Yang2020One} recently described a general procedure
that enables many geometric estimation problems involving the (robust) truncated
squared-error loss to be reformulated as polynomial optimization
problems,\footnote{This reformulation involves introducing an auxiliary binary
  indicator variable, similarly to the approach of switchable constraints
  \cite{Sunderhauf2012Switchable}.} and demonstrated empirically that the
semidefinite relaxations \cite{Lasserre2010Moments} of these problems are
frequently tight, even when a large fraction of the data (in their experiments,
greater than 50\%) are outliers. While the resulting relaxations are too large
to be solved directly using standard (interior-point) methods, they derive an
efficient algorithm for solving the dual (sums-of-squares) problems, thereby
enabling the verification of (correct) candidate solutions $\Est$ for the
original robust estimation problem in Equation
\eqref{MLE_formulation_of_SLAM}\footnote{This is analogous to the verification strategy
  employed by Carlone and colleagues
  \cite{Carlone2015Duality,Carlone2015Lagrangian} for the specific case of
  pose-graph SLAM.} Finally, they show empirically that a local optimization
strategy [based on graduated non-convexity \cite{Yang2020Graduated}] applied
directly to Equation \eqref{MLE_formulation_of_SLAM} very often succeeds in
recovering certifiably optimal solutions, despite its nonconvexity. Taken
together, these approaches provide an efficient and practically effective means
of recovering certifiably optimal robust estimators $\MLE$ for high-dimensional
problems, even in the presence of substantial outlier contamination.

\subsection{Information-Theoretic Limits of SLAM}

Having considered the computational and algorithmic challenges of the
M-estimation in Equation \eqref{MLE_formulation_of_SLAM}, we now address the
statistical properties of the estimator $\MLE$ itself. In particular: can we
develop sharp limits and/or guarantees on the statistical performance of $\MLE$
in SLAM?

As in the case of Sections \ref{certifiably_correct_SLAM_subsection} and
\ref{Certifiably_robust_subsection}, the fact that instances of Equation
\eqref{MLE_formulation_of_SLAM} are typically defined over high-dimensional
nonconvex spaces can substantially complicate the analysis;
nevertheless, recent work has identified certain spectral graph-theoretic
properties \cite{Chung1997Spectral} of the model (Figure \ref{fig:slam-example})
underlying an instance of SLAM as the key quantities controlling estimation
performance, at least in certain practically-important cases. Specifically, the
connection Laplacian $L$ [a generalization of the standard (scalar) Laplacian to
graphs with matrix-valued data assigned to their edges \cite{Singer2011Angular}]
and its spectral gap $\lambda(L)$ have emerged as objects of central importance.
For rotation averaging, Boumal et al.\ \cite{Boumal2014Cramer} showed that the
Cram\'er-Rao bound (a lower bound on the achievable covariance of any unbiased
estimator) admits a simple expression in terms of the connection Laplacian $L$.
Similarly, Khousoussi et al.\ \cite{Khosoussi2019Reliable} have recently derived
an analogous result for 2D pose-graph SLAM. These results provide simple and
sharp relations between the graphical structure of SLAM problems and the
accuracy of SLAM estimates.

Interestingly, the analyses presented by Bandeira et al.~\cite{Bandeira2016Tightness}, Rosen et al.~ \cite{Rosen2019SESync}, and Eriksson
et al.~\cite{Eriksson2018Rotation} showed that the spectral gap also plays a
central role in controlling the exactness of the semidefinite relaxations
underlying the certifiably correct estimators for the rotation averaging and
pose-graph SLAM problems; that is, the same quantity $\lambda(L)$ controls both
the statistical and the computational hardness of these estimation problems.
While much work remains to be done in this area, these early results are
strongly indicative that spectral graph-theoretic tools will have an important
role to play in designing reliable measurement networks for spatial perception.

\subsection{Open Questions and Future Research Directions}

The theoretical and algorithmic advances described in this section lay the
foundation for designing a new generation of principled, efficient, and provably
reliable estimation algorithms for SLAM. However, much work remains to realize
this potential in the form of standard technology that can be readily deployed
by practitioners. In this section, we highlight three avenues for future work
toward achieving this goal.

\subsubsection{Efficient optimization methods for certifiably correct perception}

The high computational cost of semidefinite programming remains a serious
obstacle to the development of practical certifiably correct estimation methods.
For example, all of the certifiably correct methods described in Section
\ref{certifiably_correct_SLAM_subsection} are either restricted to small-scale
problems that can be solved using standard (interior-point) semidefinite
programming techniques, or else depend on specialized, purpose-built semidefinite
optimization algorithms that are specifically tailored to each problem's
structure. While semidefinite optimization remains a very active research area
\cite{Majumdar2019Scalability}, the development of computational approaches that
are reasonably general, easy to use, and well-suited to the particular
characteristics of machine perception applications (e.g.,\ high-dimensionality,
ill-conditioning, limited computational and temporal resources, and the need for
high-precision solutions) remains an open problem. Reference
\cite{Rosen2020Scalable} reports one initial step in this direction, but much
work remains to be done.

Similarly, to date work on certifiable estimation has primarily addressed the
offline (batch) setting. The development of efficient incremental semidefinite
optimization methods---analogous to, e.g., iSAM
\cite{Kaess2008iSAM,Kaess2012iSAM2ijrr} and GTSAM \cite{Dellaert17now}---would
be extremely valuable for enabling certifiable estimation in real-time online
perception tasks.

\subsubsection{A priori performance guarantees}  

Certifiably correct perception methods approach the problem of solution
certification in a post hoc, per instance manner. This is sufficient to enable
run-time verification and monitoring to confirm that a perception system is
functioning as it should be. However, as designers and practitioners, we would
also like to have formal results that clearly delineate in advance the
circumstances under which such systems will succeed, as well as
characterizations of their expected performance as statistical estimators. While
there are some formal results that guarantee at least the existence of a noise
regime within which certifiable perception methods will succeed
\cite{Rosen2019SESync,Eriksson2018Rotation,Fan2019Efficient,Cifuentes2020Geometry},
at present there do not appear to be general, user-friendly theoretical tools
for deriving sharp bounds on the size of this regime. Results of this type would
be extremely useful in the design of measurement systems for machine perception
applications, especially in safety- and life-critical applications (e.g.,\
autonomous vehicles).

\subsubsection{Beyond point estimation}

The certifiably correct methods described in this section are all derived in the
context of Equation \eqref{MLE_formulation_of_SLAM}, i.e., in the setting of
point estimation; this is a reasonable approach whenever the underlying
likelihood or posterior probability distribution is highly concentrated around a
single mode. However, in practice it often occurs that the posterior is highly
diffuse (due to a lack of sufficiently informative measurements), or even
multimodal (as in the case, for example, of uncertain data association). In
these cases, point estimation can dramatically underestimate the actual
posterior uncertainty, even missing the existence of completely distinct but
equally plausible solutions. Blindly trusting such a result can easily cause an
erroneously overconfident belief in a completely wrong answer, potentially
endangering the safety of the overall system.

Addressing this challenge will require the development of inference methods that go
beyond simple point estimation and attempt to explicitly characterize posterior
uncertainty. This is necessary both for introspection (i.e., to enable an
autonomous agent to know what it doesn't know), and, by extension, for planning
and active perception (to enable an autonomous agent to reason about how it
could reduce its own uncertainty). The development of tractable estimation
methods that can extract this richer information while scaling gracefully to
high-dimensional problems is an important and fundamental open problem for
future research in SLAM, although References
\cite{Fourie2016Nonparametric,hsiao2019mh,bernreiter2019multiple} have proposed
initial steps along these lines.

\section{REPRESENTATION: BEYOND POINTS AND PLANES}
\label{Representation_section}

This section surveys recent advances in map and robot state representation for
SLAM. Our focus throughout is on the development of representations that go
beyond geometry alone and the corresponding problems that arise in
this domain. A major throughline of recent work in representations for SLAM is
the unification of semantic and geometric information, propelled by recent
advances in machine learning. We discuss the challenges,
opportunities, and major open questions associated with the development of joint
geometric and semantic representations; in so doing, we highlight three key
research areas, considering the influence of semantics in each: navigation in
dynamic and semi-static environments, abstraction and hierarchy, and learned
representations. In order to contextualize progress in this area, we briefly
review state-of-the-art geometric representations for SLAM.\footnote{For a
  review of progress in geometrically-grounded SLAM representations,
  particularly those used in visual SLAM, we refer readers to a survey by Cadena
  et al. \cite{cadena2016past}.}

In considering the problem of constructing a global representation from local
measurements, a fundamental question arises: What should the global
representation be? More precisely, in designing a navigating robot, a choice of
environmental representation must address which features (properties) of the
world are relevant, what data and/or models are necessary to encode those
features, and how that approach should be operationalized. Given that the
answers to many of these questions depend on the specific task at hand, it is no
surprise that numerous representations have been proposed, with no clear
universally superior choice. Moreover, while the basic problem statement for
SLAM has remained essentially unchanged for more than thirty years, the criteria
for success have changed drastically. No longer do we expect SLAM methods to
simply build geometric maps of static worlds and localize a robot. Modern SLAM
methods often must also synthesize data from heterogeneous sensors to infer
object categories, operate in dynamic and evolving environments, and support
planning. These expectations mirror the increasingly stringent requirements in
estimation for SLAM: We expect certifiability, robustness, and reliable
operation with a variety of sensors, each possessing distinct noise
characteristics.

Within the last decade, the expectation of SLAM methods to perform certain scene
understanding tasks coupled with classical geometric estimation coincided with
the recent successes (and accessibility) of machine learning methods for
perception tasks [especially in computer vision \cite{krizhevsky2012imagenet,
  redmon2016you}], driven mainly by the availability of large-scale labeled
datasets such as ImageNet \cite{deng2009imagenet}. Such advances in learning,
especially deep learning, motivated research in richer, semantic or object-level
representations for SLAM \cite{sunderhauf2018limits}. The expansion of SLAM to
include semantic perception capabilities (most prominently using vision) to
enable an embodied system to interact with the environment has recently been
referred to as spatial AI \cite{davison2018futuremapping,
  davison2019futuremappingtwo, Nicholson20online}.

Due in large part to the recent progress in learning for machine perception and
the historical success of geometric methods, the past decade has seen increased
research interest in the development of SLAM systems that estimate nonmetric
properties of the environment (e.g.,\ classifying static versus dynamic parts of
the scene), and advances in representation learning methods (e.g., deep
learning) can be used to build richer maps. The coupling of map representation
with the task (task-driven perception) is an emerging theme: We desire SLAM
systems that will build environment representations that are useful for some
task. Increasingly often, these tasks include active or interactive perception
\cite{bohg2017interactive}, intelligent exploration, manipulation, and learning
as a task in itself.

\subsection{Geometric Representations}

Sparse, landmark-based (or feature-based) representations have been used since
the earliest work on SLAM more than 30 years ago. These
representations seek to build and maintain a map of landmarks, i.e., salient,
distinct environment features which can be reliably recognized. The problem of
recognizing previously mapped features is known as data association. In the
context of visual SLAM, the SIFT \cite{lowe2004distinctive}, SURF
\cite{bay2008speeded}, and ORB \cite{rublee2011orb} features are among the most
commonly used feature descriptors. Visual SLAM methods leveraging sparse
representations, such as ORB-SLAM \cite{muratal2015orbslam, mur2017orb} and DSO
and its variants \cite{engel2017direct, wang2017stereo, gao2018ldso}, have had
tremendous success at precisely localizing a camera in three-dimensional
environments. However, the maps built by these methods, while useful for
localization, are not actionable. Geometrically, they consist of a sparsely
distributed collection of points in 3D space, rather than an explicit
characterization of free and occupied volumes and the boundaries (surfaces) that
separate them. In consequence, they are not immediately convenient for planning
collision-free paths or exposing potential routes for exploration.

In contrast, dense spatial representations attempt to build complete, albeit
approximate, descriptions of surfaces or occupied space. These descriptions may
take the form of occupancy grid maps \cite{moravec1985high}; volumetric maps (in
3D) \cite{hornung2013octomap}; meshes \cite{rosinol2019kimera}; dense point
clouds, as in LSD-SLAM \cite{engel2014lsdslam}; or truncated signed distance
function maps, as in ElasticFusion \cite{whelan2015elasticfusion} and
KinectFusion \cite{newcombe2011kinectfusion}. Meshes and other forms of surface
representations provide critical information for a system attempting to avoid
colliding with its environment, whereas point-cloud-based methods do not. In
these dense representations, the correspondence problem is addressed by
computing the most probable location within the map from which a measurement
could have been made. In many geometric maps, this is performed through a
variant of iterative closest point registration; alternatively, as with sparse
representations, loop closures can be determined through place recognition using
feature descriptors [e.g., as is done in the dense SLAM system Kintinuous
\cite{whelan2012kintinuous}].
 
The principal limitations of solely geometric representations are as follows:
\begin{enumerate}
\item Geometry alone cannot explain all potentially relevant sensory properties
  of the environment (e.g., color, tactile sensation, or weight). Thus, a
  comprehensive understanding of the environment requires reasoning capabilities
  above and beyond geometry.
\item This fact suggests that geometry alone is insufficient for more
  sophisticated forms of sensor fusion that capture not only the physical
  appearance of objects, but also interaction, sound, touch, and so on.
\item Bare geometric representations do not naturally support human-robot
  interaction. For most tasks, humans do not specify, e.g.,\ objects or
  locations in terms of numerical class labels or spatial coordinates. If robots
  aim to interact with the world alongside humans, then we need them to have at
  least some basic competency at interpreting higher-level, human-centric
  semantic descriptions.
\item Though spatial abstractions for geometric data exist, they are not
  grounded in action. Any representation suited to reasoning about a task beyond
  localization must be actionable.
\end{enumerate}

\subsection{Unifying Semantics and Geometry}

By 2015, the SLAM community had increasingly recognized the limitations of
purely geometric perception and that concurrent advances in machine learning
would enable richer, semantic representations \cite{cadena2016past,
  sunderhauf2015slam}. Research on semantics in SLAM considers the development
of more expressive map representations capable of incorporating objects and
places, and enabling higher-level autonomy. The ability to understand the world
in terms of objects and places can provide robots with a number of benefits over
traditional (dense or semidense) approaches, such as point clouds or octrees.
Semantic maps can be encoded with much smaller memory and processing footprints,
and can provide robustness against the inevitable accumulation of small errors
that can render purely geometric approaches brittle. Map representations based
on human-understandable semantic primitives can also enable better ways for
human operators to interact with autonomous systems, in more natural terms for
the human. Thus far, there is no consensus on a mathematical formalism for the
fusion of semantic information about the state of a robot and its environment
with estimates of the local scene geometry. This section concerns the progress
that has been made in establishing models for semantic information in SLAM
systems.

The prevailing approach toward incorporating semantics into navigation systems
is to treat the output of learned perception models (e.g., object detectors) as
virtual sensor measurements. As in classical work on learned sensor models, this
treatment essentially posits that some deterministic (though perhaps unknown)
function relates the latent semantic category of an object with some other
measurable physical properties, such as its appearance in a camera image, and
that this function can be approximately learned from data. Historical efforts in
robot mapping have sought to learn complex measurement models, e.g., those of
sonar sensors \cite{moravec1993learning}. However, in this prior work there
really is a (perhaps complex) relationship between the geometric structure of a
scene and the measurements made in that environment; this is an immediate
consequence of the physics of the sensing apparatus. In contrast, such a
physically-grounded relation need not hold between raw sensory data and
semantics.

Semantics often arise through affordance rather than appearance. Such
affordances may only become clear through interaction, and one cannot reason
about interaction using single-image measurement models that currently dominate
the navigation literature. Nonetheless, the recognition of objects from an a
priori known set of classes---enabled by these approaches---is undoubtedly a
useful capability for mobile robots. Recent work has focused on challenges
arising within this relatively limited scope, such as characterizing the noise
or uncertainty in the output of learned perception models, and the fact that
these models are known to fail unpredictably even in nominal, non-adversarial
operating conditions, causing drastic errors in systems that employ this
information for navigation tasks. However, incorporating discrete measurements
of object categories into the continuous geometric formulation of SLAM poses a
challenge for inference methods: Joint discrete-continuous estimation often
leads to combinatorially large state spaces, making the determination of the
most probable map difficult.

\subsubsection{Joint inference of semantics and geometry}

Given a model of object detections and classifications as the output of a
(noisy) sensor, the problem of jointly estimating the latent semantic class and
geometry of landmarks in an environment can be posed in terms of Equation
\eqref{MLE_formulation_of_SLAM} as:
\begin{equation}
  \label{semantic_SLAM}
  \hat{X}, \hat{L}, \hat{D} = \arg\max_{X, L, D} p(X, L, D \mid \Data),
\end{equation}
where $\Data$ denotes the full set of measurements (including semantic
measurements); $X$ the set of robot poses; $L$ the set of environmental
landmarks, which typically consist of some geometric information (e.g.,
position, orientation, and size) coupled with a discrete semantic label from a
known, fixed set of classes; and $D$ the set of associations between
measurements in $\Data$ and landmarks in $L$. A key observation is that
discrete-valued categorical information about objects can be naturally combined
with the already discrete inference problem of data association: Knowledge of an
object's category can help distinguish it in clutter from other objects. This
formulation unifies discrete models of semantic category, geometric estimation,
and data association; however, in addition to being nonconvex and
high-dimensional (as in the standard SLAM formulation), it now also involves
combinatorial optimization. Moreover, in committing to the use of semantics for
data association, one must cope with the errors of learned perception models.

Given this formulation, Bowman et al. \cite{bowman2017probabilistic} performed
joint optimization via expectation maximization: First fix the data association
probabilities and landmark classes and optimize the robot poses and landmark
locations (with measurements weighted by the respective probabilities of their
landmark correspondence), then fix the robot poses and landmark locations to
compute new data association probabilities and landmark classes. This approach
has the benefit of assigning soft associations to objects, gradually converging
on a locally optimal solution to the problem in Equation \eqref{semantic_SLAM}.
The data association probabilities can be computed via approximate matrix
permanent methods \cite{atanasov2014semantic}.

An alternative approach to the combinatorial inference problem of Equation
\eqref{semantic_SLAM} is to reframe the optimization over discrete variables as
one over only continuous-valued variables. In early work to this end,
S\"underhauf et al. \cite{sunderhauf2015slam} optimized probabilities of
semantic labels, which are defined over the $(K-1)$-dimensional simplex for
a $K$-class semantic labeling problem. More recently full (albeit approximate)
posterior inference has been considered by marginalizing out the discrete
variables, producing a mixture representation \cite{doherty2019multimodal}. A
similar methodology has been applied in the context of maximum a posteriori
inference to enable the use of continuous, gradient-based optimization methods
to approximately solve Equation \eqref{semantic_SLAM}
\cite{doherty2020probabilistic}. Finally, some recent work has approached the combinatorial
problem directly via multi-hypothesis tracking \cite{bernreiter2019multiple}.

\subsubsection{Semantic map representations}

Many representations for semantic navigation consist of traditional geometric
representations (such as truncated signed distance functions, occupancy grids,
or meshes) in which each element is augmented with a semantic label.
SemanticFusion \cite{mccormac2017semanticfusion} is one such dense
representation; semantic octree-based occupancy maps \cite{sengupta2015semantic}
have also been used, and Kimera \cite{rosinol2019kimera} uses meshes.

Much work has also been done in the past several years in the area of
object-level representations within SLAM. Since the early work on these
representations \cite{salas2013slam++, civera2011towards, castle2007towards},
the community has largely shifted away from a priori known object models toward
the use of learned perception models for object detection, recognition, and pose
estimation. The simplest object-centric representations treat object landmarks
as points in Euclidean space augmented with semantic labels. More recently,
representations have been developed that permit the estimation of not just the
class and position of objects, but also their orientation and extent. These
include the dual quadric formulation \cite{nicholson2018quadricslam,
  sunderhauf2017dual, ok2019robust}, which models objects as 3D ellipsoids, as
well as CubeSLAM \cite{yang2018cubeslam}, which represents objects in terms of
rectangular bounding volumes. Another recent consideration in work on
object-centric representations has been the use of learned object descriptors;
an example is the work of Sucar et al. \cite{sucar2020neural}, which captures
the shape (via the occupied volume) of objects as well as the pose of the
object.

A major theme among all of these representations is the use of some principally
geometric representation augmented with semantic class. In subsequent sections,
we discuss the incorporation of more complex semantic relationships into SLAM
representations, though the development of such models is an important
underexplored area that we revisit explicitly in Section
\ref{subsection:open_questions}.

\subsection{Beyond Static Worlds}

The vast majority of SLAM systems have assumed that the world is static. This static world assumption (i.e., that only the robot itself can change state)
has enabled great progress in SLAM, but is often violated in practice. In
consequence, most practical implementations of SLAM implicitly regard dynamic
objects as unmodeled disturbances, and rely on outlier rejection mechanisms
[e.g., RANSAC \cite{Fischler1981RANSAC}] to filter them out. More recently,
methods that attempt to explicitly discriminate between dynamic and static
objects in a visual scene have been developed in order to identify (static)
areas of the scene that are more informative for localization
\cite{bescos2018dynaslam}. Particular interest in recent research is not
restricted to the removal of dynamic elements from a scene, but also extends to
modeling their dynamics over a variety of spatial and temporal scales.

In going beyond the static world assumption, a fundamental question arises: Did
I move or did the world move? Often, without more information, a conclusion
cannot be drawn. Prevailing methods that rely on the static world assumption
break in scenarios where the entire scene moves but the robot stands still. A
key representational element missing from such systems is the ability to reason
about geometric ambiguity. Formally, this problem stems from a lack of
observability. To achieve truly robust perception, a SLAM system must be able to
reason about such ambiguous information. Furthermore, in light of making an
error of judgment, such a system should be able to revise its decision, thereby
fixing its error, and using its corrected representation to resume state
estimation. This scenario can be experienced, for example, when one is stopped in traffic
and a neighboring car begins to move: One may feel that it is oneself moving,
when in reality it is the scene that moved. Priors grounded in semantics can
help to address this observability problem: Knowledge that certain collections
of geometric features correspond to a house rather than a car may suggest
(though not ensure) that the landmark is stationary. The modeling of changing
environments poses additional, more abstract challenges. If we allow
environments or the objects within them to change over time, then the
recognition of familiar places and identities of objects within becomes
nontrivial. Like the ship of Theseus, whose pieces are gradually replaced until
none of its original components exist, it is difficult or impossible in dynamic
worlds to determine unambiguous object or scene identities. Consequently,
solutions to SLAM in dynamic environments generally rely on some assumptions
about the nature of the possible dynamics in order to make the problem
tractable.

\subsubsection{Dynamic environments}

Operationally, highly dynamic environments consist of those in which any motion
not due to the robot occurs on a timescale that makes it directly observable
(pedestrians, cars, etc.). This is the simplest and most commonly studied form of dynamic
SLAM. The direct observability of these motions reduces the problem of dynamic
SLAM to a front-end classification and filtering problem: Dynamic components of
the scene can be identified and modeled locally without becoming part of the
global environmental representation [as in the work of Wang et al.
\cite{wang2007simultaneous}]. Recent methods such as those described in
References \cite{schorghuber2019slamantic, zhang2020vdo, henein2020dynamic}
leverage semantics to inform dynamic SLAM systems, particularly in deciding
which features correspond to objects that are likely to be moving. An emerging
theme in this area is the estimation of the velocity of classified objects in
the scene.

\subsubsection{Semistatic environments}

Semistatic environments are those in which environmental change happens on a
timescale that is observable only by repeated revisitation or observation of a
given location. This can include, for example, furniture moving around or
seasonal variation. The key challenge associated with modeling semistatic
environments is that, operationally, learning and reasoning about these kinds of
environmental changes requires some implementation of environmental memory. This
difficulty is in large part representational: What does it mean for a map to
change over time? How can we build updatable maps, and how should we think about
recording the evolution of environments over time once changes are permitted?
Numerous approaches to this have been proposed. Rosen et al.
\cite{rosen2016towards}, for example, developed a Bayesian framework for
recursive estimation of the persistence of each feature in an environment.
Krajn\'ik et al. \cite{krajnik2017fremen} applied Fourier analysis to model the
frequencies at which features in the environment may change. Bore et al.
\cite{bore2018detection} made use of particle filters to learn the temporal
dynamics of an arbitrary number of objects with a priori unknown feature
correspondence. Zeng et al. \cite{zeng2020semantic} considered a semantic
linking map representation that captures probabilistic spatial couplings between
landmarks; this captures the intuition that in a semistatic environment,
contextual semantic relationships between observed objects can be used to
facilitate object search (e.g., to find cups, we might first look for tables or
cabinets). Halodov\'a et al. \cite{halodova2019predictive}, Berrio et al.
\cite{berrio2019updating}, and Pannen et al. \cite{pannen2020how} each
considered the problem of managing changes to a map representation (though they
did not explicitly attempt to track dynamic objects over time). The problem of
change detection and map modification is especially important for
safety-critical autonomous vehicle systems that rely on high-resolution 3D maps
for navigation.

\subsubsection{Deformable environments}

State estimation in deformable environments, or environments with deformable
objects, remains a particularly challenging and underexplored research area in
SLAM. A major milestone on this front was DynamicFusion
\cite{newcombe2015dynamicfusion}, which demonstrated dense reconstruction of
deformable objects using an RGB-D camera. More recent work has investigated the
 observability of SLAM in deformable
environments, in particular with application to mapping the interior of the
human heart \cite{song2019observable}.

\subsection{Abstraction and Hierarchy in Spatial Representations}

Abstraction and hierarchy in spatial environment models have greatly improved
scalability, efficiency, and generalization of SLAM methods. Early work on the
spatial-semantic hierarchy \cite{kuipers2000spatial} outlined key ideas related
to the construction and maintenance of a cognitive map of the environment. From
a purely geometric standpoint, spatial partitioning algorithms enabled the
scaling of dense representations of occupancy \cite{hornung2013octomap}.

\subsubsection{Topological models}

Environment topology plays a major role in decision-making. For
navigation, the (topological) decision to go left or right at a fork in the road
can have a far more significant impact on the time to reach a destination than
decisions about the specific (metric) motion plan. Topological maps describe an
environment at this level of abstraction, i.e., at the level of connectivity. In
the context of SLAM, topological maps provide a very compact representation that
can be used for such coarse (but significant) navigation decisions, as well as
localization.\footnote{For a review of topological methods prior to 2016, we
  refer readers to \cite{lowry2016visual}.}

In particular, topological feature graphs \cite{mu2016information}, whose
vertices represent geometric features and whose edges represent obstacles, have
recently been used to support information-theoretic exploration, where they
provide a compact description of occupied space. Stein et al.
\cite{stein2020enabling} proposed a polygonal map representation from which
topological navigation decisions can be obtained, enabling a broad class of
learning-aided planning tasks. Neural topological SLAM \cite{chaplot2020neural}
incrementally constructs a topological map (graph) in which vertices represent
physical locations (identified with a set of features extracted from panoramic
images via a neural network-based encoder) and edges represent
connectivity/traversability between locations; this representation is used to
support learning policies to navigate toward a goal, specified as an image taken
at the target location.

\begin{figure}
  \centering
  \includegraphics[width=.75\textwidth]{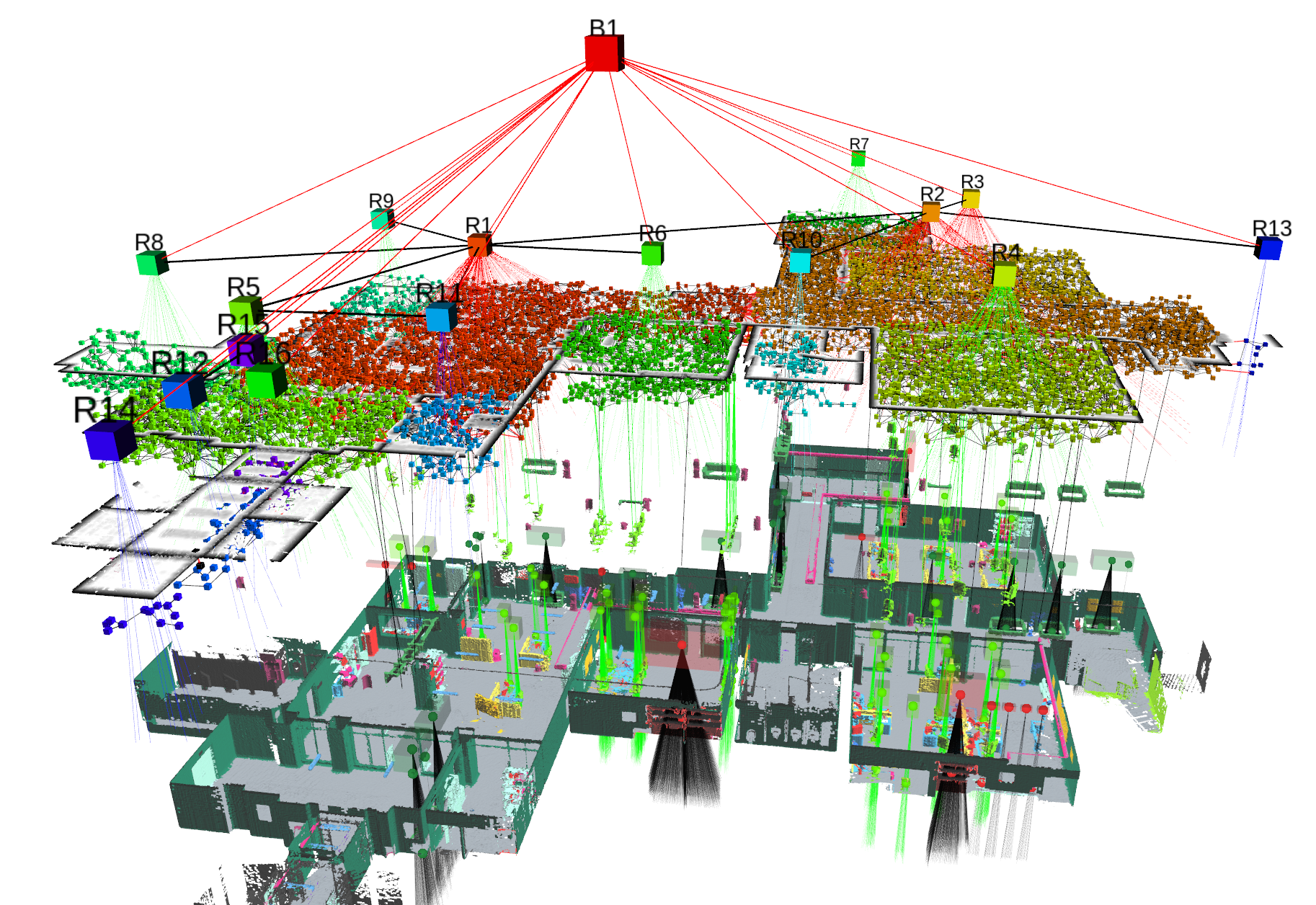}
  \caption{3D dynamic scene graphs \cite{rosinol20203d}. These are a recent
    application of scene graphs (previously common in the computer graphics
    community) to the SLAM problem and provide a substantial step toward linking
    scene understanding and spatial perception methods. Figure courtesy of A.\
    Rosinol.} \label{fig:reps}
\end{figure}

\subsubsection{Scene graphs}

Representations that synthesize spatial and semantic information in hierarchies
incorporating both metric and topological properties of the environment have
been of very recent interest. Hierarchical models have the benefit of
partitioning space at a variety of levels of abstraction, enabling efficient
reasoning over large spatial scales. In particular, 3D scene graph models
\cite{armeni20193d, rosinol20203d} present a promising representational
direction toward capturing object-level semantics, environment dynamics, and
multiple spatial and semantic layers of abstraction (from the connectedness of
unoccupied space, to rooms and buildings, and beyond). Scene graphs model the
environment in terms of a directed graph where nodes can be entities such as
objects or places and edges represent relationships between entities (depicted
in Figure \ref{fig:reps}). The relationships modeled by a scene graph may be
spatial or logical. Scene graph representations have also been used to learn
physical descriptions of scenes in an unsupervised fashion from visual input
alone \cite{bear2020learning}, though this approach has yet to be applied
directly in a robotics context.

\subsection{Learning-Centered and Learned Representations}
\label{subsection:learning}

Many efforts thus far in semantic SLAM leverage learned perception models but
opt for relatively straightforward environmental representations; these
typically consist of some combination of classical geometric measurements and
categorical output from a learned model, such as an object detector (e.g.,\
\cite{redmon2016you}). Such perception models are typically trained in an
offline setting, using data that may not accurately reflect the conditions in
which a robot is actually deployed. We may ask, then, whether it possible to
learn or refine perception models during robot navigation. In particular,
knowledge of the scene geometry and robot motion during navigation motivates
self-supervision of learned perception models (e.g., \cite{detone2017toward,
  detone2018self}). We must address, in such settings, precisely what sorts of
environment representations would permit these types of learning. Taking this a
step further, we might consider learning parts of the representation itself, the
logical extreme of which is end-to-end learning of navigation (as in
\cite{bojarski2016end}).

\subsubsection{SLAM for self-supervised learning}

Work on SLAM-aware perception methods originated in the desire to leverage
global spatial structure when recognizing objects in video streams
\cite{pillai2015monocular}. Since then, these capabilities have been extended to
bootstrap supervision for a variety of learned perception models
\cite{Pillai17iros, Pillai17irosworkshop, detone2017toward, detone2018self}. The
central idea of these approaches is that a global (typically geometric)
representation of the environment can be used as a supervisory signal for
training a variety of models relating the motion of a camera and the scene
geometry (e.g., visual odometry models). The ability to combine spatial and
temporal information in order to improve learning is a unique feature of
embodied spatial AI systems that undoubtedly warrants further attention.

\subsubsection{Learning geometric representations}

Coarse geometric models often fail to capture the precise volumes of objects,
while dense representations require substantial memory to store a map. SceneCode
\cite{zhi2019scenecode} and NodeSLAM \cite{sucar2020neural} incorporate learned
intermediate representations for objects. From these compressed representations,
the object geometry can be recovered. More generally, DeepFactors
\cite{czarnowski2020deepfactors} incorporate compact representations of depth
images toward the same goal.

Given the popularity of gradient-based methods for optimization (particularly
backpropagation), there is great potential utility for differentiable SLAM
representations. Recent work to this end \cite{jatavallabhula2019gradslam} aims
to enable backpropagation through traditional SLAM systems in order to
seamlessly integrate with learning models such as neural networks.

\subsubsection{Learning to navigate}

Recently, several models have considered end-to-end learning for navigation.
Broadly, these methods seek to learn functions mapping directly from sensor
inputs (or a history of sensor measurements) to actions (e.g.,
\cite{zhang2017neural, mirowski2018learning}). A major consideration for these
methods is how to structure the learning problem. Zhang et al.
\cite{zhang2017neural} used an agent with external memory representing an
occupancy map. More recent work \cite{chaplot2020neural} has considered
specifically structuring the learning problem to use a topological
representation of the environment.

\subsection{Open Questions and Future Research Directions}
\label{subsection:open_questions}

While the previous decade has brought great progress in the representational
power of modern robotic mapping and localization systems, a number of
fundamental issues remain open. In this section, we highlight three key avenues
for future work: novel representations for sensor fusion, hierarchical
abstractions for learning and attention, and the problem of identifying suitable
concepts for grounding semantic models.

\subsubsection{Novel representations for sensor fusion}

While much progress has been made in recent years on the topic of semantic SLAM,
most representations rely on the availability of high signal-to-noise ratio
sensors that provide an abundance of accurate geometric measurements, such as
RGB-D cameras and lidar. How can we do more with less? We would like to develop
robots that process rich, multimodal sensory information from measurements that
may individually only partially describe objects or scenes. Incorporating novel
sensors such as event cameras [see, e.g., the recent survey by Gallego et al.
\cite{gallego2019event}], light-field cameras, and tactile sensing together with
more traditional sensors such as cameras, lidar, inertial measurements, and
sonar is a key area for future work. Similarly, the graphical optimization-based
estimation frameworks that at present are commonly used in robotic state
estimation provide a convenient and versatile language for describing sensor
fusion problems, but they typically depend on having a relatively
well-characterized model for each of the deployed sensors. Natural organisms, as
well as machines, encounter physical changes to their sensors and configurations
as time goes on: A camera may move slightly on an autonomous vehicle, lens
distortion may occur, and so on. Systems that perform long-term sensor fusion
must in some sense be adaptable. There is still ample room for fundamental
contributions in many of these areas.

\subsubsection{Hierarchical models for learning, navigation, and planning}

Hierarchical, flexible models that abstract the minutiae of scene geometry in
favor of higher-level concepts are needed for robust, scalable long-term
mapping. However, the design of these abstractions raises several issues,
including to what extent they should be learned or grounded in human-centric
concepts, and how they can be structured to accommodate multiple spatial and
temporal scales (to support large-scale operation).

\paragraph{To learn or not to learn?}

Thousands of years of human-made spatial-semantic abstractions have become
embedded into our world. Given these prior abstractions, two natural questions
arise: what should be learned from data, and what should be enforced as a prior?
In particular, should a robot be explicitly programmed with human-centric
abstractions, or should it attempt to learn its own abstractions through
experience? The answer to this question will naturally be task-dependent, and
the twin problems of developing both models of human semantics for robot use and
online robot semantics (grounded in the actions a robot can perform) remain
largely open.

\paragraph{Flexible representations and attention}

What is the spatial representation for a robot that can get on a plane and fly
from Boston to London? It may not matter explicitly where it is in
Earth-centered coordinates while on an airplane: Only the local coordinates of
the robot matter. Mixing topological constraints with metric constraints would
be greatly beneficial in such situations. While this topic was initially
explored by Sibley et al. \cite{sibley2010planes}, this is largely an open
research area.

For the scalability of representations, another open question is what the robot
should remember. Metric geometric models grow unboundedly in the space needed
for navigation; however, it seems intuitively clear that not all metric
information needs to be immediately accessible. Neither a human nor a robot
needs to reason about the dense geometry of their home kitchen in order to
navigate from one location to another while traveling in another country.
Developing models that can pull relevant information from long-term memory into
short-term memory for navigation tasks is another promising area for future
work.

\subsubsection{Where do semantics come from?}

In comparison to work using semantics as a sensor, there has been comparatively
little work on semantics as they arise from the compositions of other entities
in a map, or from the history of a robot's interactions with (and observations
of) the environment. Groupings of geometric features (point clouds, lines, and
planes), or local unoccupied regions (i.e.,\ places), can each be associated
with semantic phenomena in the environment. It may be necessary to build up a
semantic model of the environment, rather than capture semantics in single
measurements, in order to develop more intelligent robot systems. The following
passage by Marr \cite[p. 36]{marr1982vision} is relevant in this context:

\begin{quote}
  Finally, one has to come to terms with cold reality. Desirable as it may be to
  have vision deliver a completely invariant shape description from an image
  (whatever that may mean in detail), it is almost certainly impossible in only
  one step. We can only do what is possible and proceed from there towards what
  is desirable. Thus we arrive at the idea of a sequence of representations,
  starting with descriptions that could be obtained straight from an image but
  that are carefully designed to facilitate the subsequent recovery of gradually
  more objective, physical properties about an object shape.
\end{quote}

For a number of reasons, scenes may contain semantic information not captured
directly by individual sensor measurements. For example, the geometry of an
object larger in extent than can be captured by a single image may be relevant,
or the semantic relevance of an object may arise through interaction. More
generally, some semantic properties of interest may have no physical grounding
whatsoever; a natural example of such a property is ownership, which is
completely divorced from an object's physical makeup. Finally, semantics are
often contextual: different properties or categorizations of objects may be
important depending on the specific task at hand. There is a great need for
flexible models that can describe these aspects of the world in order to build
versatile robots that can robustly perform a variety of tasks.


\section{DISCUSSION AND FUTURE PERSPECTIVES}
\label{Discussion}

Looking forward, we see three key challenges for guiding future research in SLAM:
\begin{enumerate}
\item {\bf Long-term autonomy:} Can we improve the robustness of SLAM systems to
  enable the reliable, persistent and independent operation of robots in the
  real world?

\item {\bf Lifelong map learning:} Can we create systems that continually
  improve their mapping performance, despite (and perhaps even leveraging) the
  constant evolution of real-world environments, thereby enabling persistent
  deployment?

\item {\bf SLAM and deep learning:} Can we capitalize on the promise of recent
  breakthroughs in deep learning~\cite{ACM19online} and new semantic
  representations, while retaining the desirable properties of traditional
  model-based state estimation methods, including recently developed robust
  estimation algorithms?
\end{enumerate}

In our view, long-term autonomy (i.e., the capacity of a robotic system to
operate reliably, for extended periods of time, without human supervision or
intervention) is an important measure of performance for evaluating future
research in SLAM. Current state-of-the-art methods have largely solved the
problems of spatially- and temporally-bounded deployment; what remains to
address are the myriad infrequently-encountered failure cases that arise in
extended, real-world operation. By nature, these may be difficult to capture in
laboratory settings via small-scale experiments, or in the standard datasets
that have traditionally been employed for empirical SLAM evaluation. We expect
that the certifiable and robust SLAM inference methods described in Section
\ref{Geometric_estimation_section} will have a particularly prominent role to
play in addressing such long-tail failure modes, as their explicitly delineated
operational assumptions and run-time verification enable potentially hazardous
circumstances to be identified before they are encountered in operation.

\begin{figure}
\centering
\includegraphics[width=0.8\textwidth]{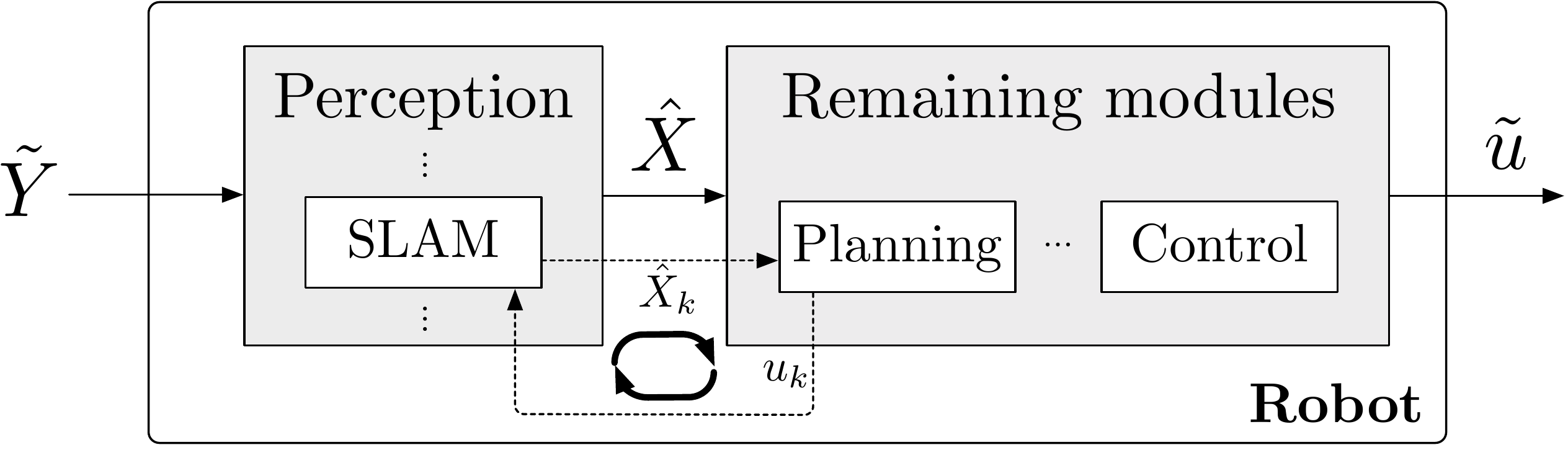}
\caption[Overview of the entire robotic system]{Block diagram of a robotic
  system, including perception and SLAM modules. Solid lines represent the basic
  flow of information through the perception subsystem: Sensor measurements
  $\Data$ are received and processed, and a state estimate $\Est$ is
  communicated to the rest of the system. Active perception
  \cite{bajcsy1988active} involves a tighter interconnection between planning
  and perception (\emph{dashed loop}): Here, the planner proactively chooses its
  controls $u_k$ to reduce the uncertainty in the predicted state estimate
  $\Est_k$ that is expected after $u_k$ is applied.}
\label{fig:robot-sys}
\end{figure}

Persistent operation will also require robots to evolve beyond the classical
snapshot version of the SLAM problem and embrace a long-term existence via
lifelong learning. The environment (and perhaps features of the robot itself)
will change over time, and new inferential and representational approaches are
required that are similarly adaptive. In particular, an important novel
challenge in this regime is the need to explicitly account for, and actively
manage, the uncertainty inherent in an ever-changing world. For long-term
operation, inference and representation must be combined with planning and
control to enable active, task-directed perception; these capabilities will
provide autonomous agents the means to introspect (i.e., to monitor their own
state of knowledge), equipping them to identify and seek out the information
needed to reduce their own uncertainty (Figure \ref{fig:robot-sys}). Recalling
Bajcsy's \cite{bajcsy1988active} famous adage, intelligent agents do not just
see, they look.

Regarding learning and adaptation, the integration of SLAM with deep learning
specifically will be another key research area over the next decade. Indeed,
end-to-end, data-driven machine learning techniques for SLAM are already
starting to enter the
literature~\cite{Pillai17iros,Pillai17irosworkshop,detone2017toward,jatavallabhula2019gradslam}.
There is ample opportunity to develop novel deep learning systems that are
specifically adapted to the unique features of robotic perception; these include
rich, temporally-coherent streams of sensory data available to robots, novel
sensing modalities and data types beyond classical vision (e.g.,\ direct 3D
perception via lidar, event-based cameras, and tactile sensing), and the ability
to close the loop around perception via active sensing. The community is also in
need of larger and more varied data sets, tailored to the problem of SLAM, to
more thoroughly investigate the potential of these approaches.

After 30 years of progress, the problem of constructing global representations
from local measurements continues to inspire significant and fundamental
advances in SLAM. Recent work has seen great strides in classical state
estimation (Section \ref{Geometric_estimation_section}), demonstrating both the
theoretical and empirical effectiveness of certifiable and robust inference
methods. However, these methods derive much of their power from being built atop
well-characterized (typically geometric) models; such strong hypotheses may not
always be realistic, especially in moving beyond short-term operation and
representations grounded in simple geometric primitives. Conversely, it is the
flexibility and representational power (Section \ref{Representation_section}) of
learning systems that affords autonomous robots the capacity to build richer
environmental models and adapt through experience. At present these systems are
often trained and deployed in a black-box end-to-end manner; this may make the
learned representations difficult to interpret, and hence difficult to integrate
within larger-scale autonomous systems (see Figure \ref{fig:robot-sys}). In the
future, we envision SLAM systems that are built as a synthesis of these
approaches, applying (narrowly-scoped) learning in those parts of the system
where it is required, but integrated within a classical model-based Bayesian
state estimation framework (Figure \ref{fig:slam-example}) that enables us to
take advantage of principled, highly-developed (e.g.,\ certifiable and robust)
inference methods, introspection, and active perception. Such a synthesis could
achieve the best of both worlds, equipping robotic agents with both the
robustness and the adaptability necessary to achieve truly autonomous persistent
operation.


\section*{DISCLOSURE STATEMENT}
The authors are not aware of any affiliations, memberships, funding, or
financial holdings that might be perceived as affecting the objectivity of this
review.

\section*{ACKNOWLEDGMENTS}
This work was supported by Office of Naval Research Multidisciplinary University
Research Initiative grant N00014-19-1-2571 and Office of Naval Research grant
N00014-18-1-2832.

\bibliographystyle{ar-style3} 
\bibliography{refs2020}
  
\end{document}